\documentclass{article}

\usepackage[final]{nips2016}

\usepackage[utf8]{inputenc} 
\usepackage[T1]{fontenc}    
\usepackage{booktabs}       
\usepackage{amsfonts}       
\usepackage{nicefrac}       
\usepackage{microtype}      

\usepackage{amsmath}
\usepackage{amssymb}
\usepackage{amsthm} 
\usepackage{subcaption}
\usepackage{graphicx}
\usepackage{multirow}
\usepackage{paralist}
\usepackage{caption}

\newtheorem{hypothesis}{Hypothesis}

\newtheorem{lemma}{Lemma}

\title{Cognitive Discriminative Mappings for\\ Rapid Learning}

\author{
  Wen-Chieh Fang\\
  \And
  Yi-ting Chiang\\
}

\begin{document}

\maketitle

\begin{abstract}
  Humans can learn concepts or recognize items from just a handful of examples, while machines require many more samples to perform the same task.
In this paper, we build a computational model to investigate the possibility of this kind of rapid learning.
The proposed method aims to improve the learning task of input from sensory memory by leveraging the information retrieved from long-term memory.
We present a simple and intuitive technique called cognitive discriminative mappings (CDM) to explore the cognitive problem.
First, CDM separates and clusters the data instances retrieved from long-term memory into distinct classes with a discrimination method in working memory when a sensory input triggers the algorithm. 
CDM then maps each sensory data instance to be as close as possible to the median point of the data group with the same class. 
The experimental results demonstrate that the CDM approach is effective for learning the discriminative features of supervised classifications with few training sensory input instances. 
\end{abstract}

\section{Introduction}

Scientists have interest in understanding the relations between all the levels that describe what a brain does.
Adolphs listed the top $23$ unsolved problems in neuroscience (including three "meta" issues)~\cite{Adolphs:15Unsolved}. 
Two important questions, "How does sensory transduction work?" and "How does learning and memory work?", are closely connected~\cite{Adolphs:15Unsolved}. 
Without sensory transduction that converts a sensory stimulus from one form to another, the brain cannot integrate and process the sensory input information with information stored in the memory.    

People can learn a new concept or recognize an item from just a handful of examples, while state-of-the-art machine learning algorithms typically require tens or hundreds of examples to perform with similar accuracy~\cite{LakeEtal:15Human-level}.
Scientists have long suspected that this type of "one-shot learning," or rapid learning, involves a different mechanism in the brain than gradual learning~\cite{Weaver:15How}.
We believe the underlying mechanism contains a complex information processing procedure.

Humans have five main senses: sight, hearing, taste, smell, and touch~\cite{SternbergSternberg:11Cognitive_Book}.
The human brain can combine data from sensory memory (SM) and prior experiences retrieved from long-term memory (LTM) into a single phenomenal experience.
\textit{Feature integration theory} tackles the question of how humans perceive individual features as part of the same object by proposing a two-stage process: preattentive processing and focused attention processing \cite{Treisman:86Features}~\cite{Treisman:99Solutions}. 
The basic idea is that objects are analyzed into features and that attention is necessary to combine these features to create perceptions of an object.
We apply this theory to our model to describe how sensory input data are processed in SM.    
In addition, people can retrieve prior experiences encoded in LTM. 
We believe that this procedure is one of the reasons why humans learn new concepts quickly. 
Therefore, we simulate this procedure in our model.

In this paper, we model rapid learning as a prediction task with feature transformation and augmentation procedures in the working memory.
The rapid learning problem aims at achieving high prediction performance when training a learning system with limited data drawn from SM by leveraging a relatively large amount of data retrieved from LTM.

In Section~\ref{sec:problem}, we formally define the problem and introduce the framework of our approach.
Then we present our approach, including the learning of discriminative features and the augmentation techniques in Section~\ref{sec:solution}.   
We proceed by evaluating our approach on benchmark data sets and discuss the results in Section~\ref{sec:experiments}.
Finally in Section~\ref{sec:conclusion}, we conclude the paper and offer suggestions for future work. 

\section{Problem Definition}
\label{sec:problem}
Assuming that we have a labeled data set of $N_{\mathcal{L}}$ points: $\{ (\mathbf{x}_i, l_i) \}_{i=1}^{N_{\mathcal{L}}}$.
The class $l_i$ of each data instance $\mathbf{x}_i \in \mathbb{R}^m$ is in class set $C_{\mathcal{L}}$.
We denote this data set as LTM data set $D_{\mathcal{L}}$.
Let there be another data set of interest. 
We denote it as SM data set $D_{\mathcal{S}}$, which has few labeled data instances $\{ (\mathbf{y}_i, l_i) \}_{i=1}^{N_{\mathcal{S}}}$.
Each instance $\mathbf{y}_i \in \mathbb{R}^n$.
The class $l_i$ is in class set $C_{\mathcal{S}}$.
The class set $C_{\mathcal{S}}$ is assumed to be equal to $C_{\mathcal{L}}$.
The LTM and SM data sets are in different feature spaces.
In most cases, the number of dimensions of the two input data sets are also different.

Our goal is to build a classifier by making use of the small data $\{ (\mathbf{y}_i, l_i) \}_{i=1}^{N_{\mathcal{S}}}$ from SM and the relatively large data $\{ (\mathbf{x}_i, l_i) \}_{i=1}^{N_{\mathcal{L}}}$ from LTM.
The classifier can be applied to new incoming data in SM and performs equally or even better than the classifier trained on data from SM.
Note that there is no data instance without knowing its class from SM during the training phase.



According to the definition in~\cite{PanYang:10Survey}, given two domains, LTM domain  $\mathcal{D}_{\mathcal{L}} = \{\mathcal{X}, P(X)\}$ and SM domain $\mathcal{D}_{\mathcal{S}} = \{\mathcal{Y}, P(Y)\}$, where $\mathcal{X}$, $\mathcal{Y}$ are feature spaces and $P(X)$, $P(Y)$ are marginal probability distributions, $X = \{ \mathbf{x}_i \}_{i=1}^{N_{\mathcal{L}}} \in \mathcal{X}$ and $Y = \{ \mathbf{y}_i \}_{i=1}^{N_{\mathcal{S}}} \in \mathcal{Y}$.  
If $\mathcal{D}_{\mathcal{L}} \neq \mathcal{D}_{\mathcal{S}}$, this implies that either $\mathcal{X} \neq \mathcal{Y}$ or $P(X) \neq P(Y)$.  
In this paper, we focus on the situation of $\mathcal{X} \neq \mathcal{Y}$ and $C_{\mathcal{L}} = C_{\mathcal{S}}$.

We give two definitions in our problem:
\begin{enumerate}
\item \textit{A discriminative cluster is a group of objects that share the same class label in a metric space.}
\item \textit{The radius of a cluster is the maximum distance between all the points and the median point.}
\end{enumerate}

We also give some assumptions:
\begin{enumerate}
 \item Data from SM and LTM exist for each class. Therefore our approach is a kind of supervised method.
 \item The data from SM and LTM may share \textit{no} co-occurrence features. We do not rely on co-occurrence features to train the model; therefore, our approach can extend well to heterogeneous domains. 
 \item No instance is shared across domains.
 \item The relationship between the LTM domain and the SM domain is not given. We only know that the two domains have the same classes in common.
 \item The new incoming data from SM for evaluation are unseen during the training phase.
\end{enumerate}

\subsection{Lemma}
\begin{lemma} 
A set of clusters $\mathrm{C}$ is said to be pairwise disjoint if and only if for every $\mathrm{C}_{\mathcal{L}}, \mathrm{C}_{\mathcal{S}} \in \mathrm{C}$, let $r_{\mathrm{C}_{\mathcal{L}}}$ and $r_{\mathrm{C}_{\mathcal{S}}}$ be the radii of $\mathrm{C}_{\mathcal{L}}$ and $\mathrm{C}_{\mathcal{S}}$ respectively; then 

\begin{equation}
d(\mathrm{C}_{\mathcal{L}}, \mathrm{C}_{\mathcal{S}}) > r_{\mathrm{C}_{\mathcal{L}}} + r_{\mathrm{C}_{\mathcal{S}}},
\end{equation}

where $d(\mathrm{C}_{\mathcal{L}}, \mathrm{C}_{\mathcal{S}})$ is the distance between the median points of the two clusters $\mathrm{C}_{\mathcal{L}}$ and $\mathrm{C}_{\mathcal{S}}$.
\end{lemma}

\subsection{Hypothesis}
We present the following main theoretical hypothesis.
We believe the proposed hypothesis provides a promising theoretical base for us to develop the algorithm.

\begin{hypothesis}
Given that there exists an $\emph{SM}$ domain $\mathcal{D}_{\mathcal{S}}$,
a sample $D_{\mathcal{S}}$ is from $\mathcal{D}_{\mathcal{S}}$.
If we can find a sample set $D_{\mathcal{L}}$ in the \emph{LTM} domain $\mathcal{D}_{\mathcal{L}}$, and two mapping functions $f$ and $g$ are such that the following conditions are satisfied:

\begin{enumerate}
 \item Each data instance $\mathbf{x}_i \in D_{\mathcal{L}}$ with class $l$ and each data instance $\mathbf{y}_j \in D_{\mathcal{S}}$ with same class $l$ are mapped into a common discriminative cluster corresponding to the same class $l$ in a new space.
 \item These discriminative clusters are pairwise disjoint. 
\end{enumerate}

Then, there exists at least a hypothesis $h \in H$ ($H$ is a family of hypotheses in $\mathcal{D}_{\mathcal{S}}$) that the empirical error rate  

\begin{equation}
\hat{\epsilon}_{\mathcal{U} \cup \mathcal{V}}(h) \leq \hat{\epsilon}_{\mathcal{V}}(h)
\end{equation}

where $\mathcal{U}$ is the projected \emph{LTM} sample set in the new space, and $\mathcal{V}$ is the projected \emph{SM} sample set in the new space. 
In other words, the samples in $\mathcal{U}$ help to reduce the empirical error rate.
\end{hypothesis}

\section{Proposed Solution}
\label{sec:solution}
\subsection{Main Idea}

Because the data from LTM and SM are in different feature spaces, it is desirable to find a common invariant feature space in which all data can be directly compared.
Inspired by the \textit{min-max principle} \cite{HoiEtal:08Semi-supervised} and \textit{class-based constraints}~\cite{SaenkoEtal:10Adapting}, we apply two transformations to map the LTM and SM data in order, to satisfy class constraints between the transformed points.

In order to learn two transformation functions $f$ and $g$ for the SM data classification, we first define two variables $\psi_S$ and $\psi_D$ as follows: 

{
\scriptsize 
\begin{equation}
\begin{aligned}
\label{eqn:sum_of_distance}
\psi_S (f, g) = \sum_{i \in \text{LTM}, j \in \text{SM}, l_i = l_j} \mathsf{d}_{\Omega}(f(\mathbf{x}_i), g(\mathbf{y}_j)) + \sum_{i, j \in \text{LTM}, l_i = l_j} \mathsf{d}_{\Omega}(f(\mathbf{x}_i), f(\mathbf{x}_j)) + \sum_{i, j \in \text{SM}, l_i = l_j} \mathsf{d}_{\Omega}(g(\mathbf{y}_i), g(\mathbf{y}_j)) \\
\psi_D (f, g) = \sum_{i \in \text{LTM}, j \in \text{SM}, l_i \neq l_j} \mathsf{d}_{\Omega}(f(\mathbf{x}_i), g(\mathbf{y}_j)) + \sum_{i, j \in \text{LTM}, l_i \neq l_j} \mathsf{d}_{\Omega}(f(\mathbf{x}_i), f(\mathbf{x}_j)) + \sum_{i, j \in \text{SM}, l_i \neq l_j} \mathsf{d}_{\Omega}(g(\mathbf{y}_i), g(\mathbf{y}_j)) \\
\end{aligned}
\end{equation}  
}
 
Here, $\mathsf{d}_{\Omega}(\cdot, \cdot)$ is the distance function defined in the common space $\Omega$.
$\psi_S$ sums the distance between the transformed points from LTM and SM with the \textit{same} class, while $\psi_D$ indicates the sum of the distance between the projected instances from LTM and SM with \textit{different} classes.

Then, we posit that there exist two bounds $\mathsf{u}$ and $\mathsf{l}$ such that two inequations are necessary for the generation of pairwise disjoint discriminative clusters:

\begin{equation}
\begin{aligned}
\label{constraints:dist}
\psi_S (f, g) &\leq \mathsf{u} \\
\psi_D (f, g) &\geq \mathsf{l} \\
\end{aligned}
\end{equation}
where $\mathsf{u}$ and $\mathsf{l}$ are the upper and lower bounds chosen so that $\psi_S$ should be small, and $\psi_D$ should be large. 
Given the LTM data, SM data, and the class constraints in advance, if we can learn two transformation functions $f$ and $g$ from the data to satisfy inequations~\ref{constraints:dist}, we are able to use LTM data to help improve prediction performance.


\subsection{Our Solution}




In this paper, we present a linear version.
We represent $f$ and $g$ as two \textit{linear} transformation functions, namely $\mathbf{P}$ and $\mathbf{Q}$.
That is, $f(\mathbf{x}_i) = \mathbf{P}\mathbf{x}_i$ and $g(\mathbf{y}_j) = \mathbf{Q}\mathbf{y}_j$.

\begin{figure}[h]
\centering
\includegraphics[width=0.6\linewidth]{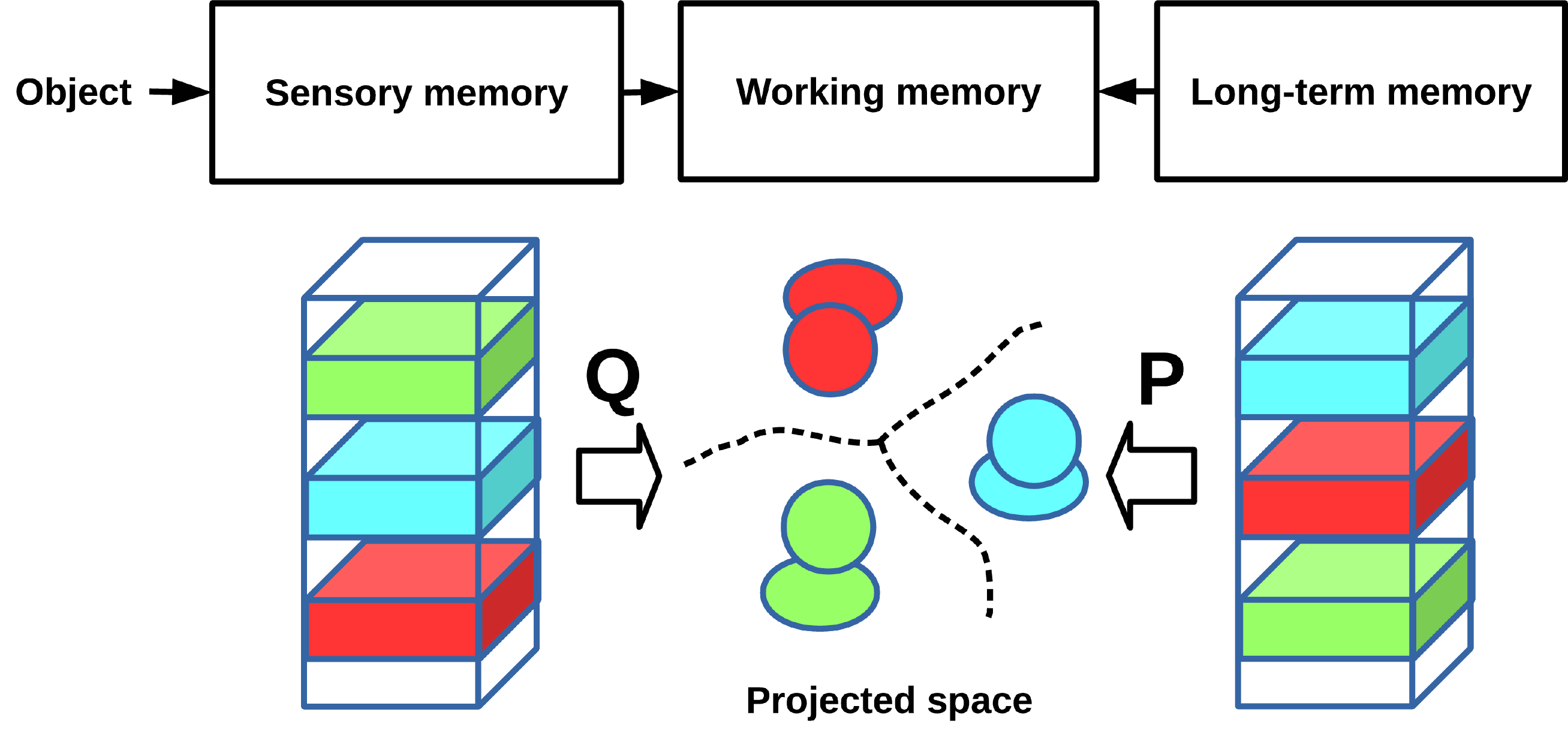}
\caption{
Illustration of Cognitive Discriminative Mappings, in which different colors represent different classes. The broken black lines represent the class boundaries. Circles represent the grouped instances from LTM while ovals represent the projected instances from SM. If the class boundaries can discriminatively separate instances into different classes, low classification errors will be expected.
}
\label{fig:model}
\end{figure}

We present a computational mappings model to make use of both data from LTM and SM.
At first, sensory data instances are being taken in by sensory receptors and kept in SM. 
When the sensory data instances are stored in SM long enough, they are transferred to working memory.
In the new working memory space, these transformed data are populated into clusters according to their classes.   
Fig.~\ref{fig:model} illustrates the main idea of the proposed model.

We can estimate the two linear transformations $\mathbf{P}$ and $\mathbf{Q}$ by first deciding one transformation.
Most of the time, there are many more related samples from LTM than there are sensory cases in SM.
Consequently, we use mapping function $\mathbf{P}: \mathbf{X} \rightarrow \mathbf{U}$ to project the LTM data $\mathbf{X} \in \mathbb{R}^{N_{\mathcal{L}} \times m}$ into $\mathbf{U} \in \mathbb{R}^{N_{\mathcal{L}} \times d}$ in a latent space $\Omega$ in working memory.
Therefore $\mathbf{P} \in \mathbb{R}^{d \times m}$.
The latent space $\Omega$ is assumed to be a $d$-dimensional Riemannian manifold. 

We present three different approaches in the retrieval phase.
The first approach is where the $c$ geometric medians of clusters corresponding to the classes are fixed and predefined in the latent space. 
The data in LTM are mapped into these locations by a transformation matrix $\mathbf{P}$ and form discriminative clusters simultaneously.

In the second approach, we apply Linear discriminant analysis (LDA)~\cite{Fukunaga:90Introduction} to characterize or separate the classes of instances.
We use the transformation matrix $\mathbf{P}$ provided by LDA to project original instances onto a latent space, and a new feature set is generated. 

In the third approach, a linear transformation $\mathbf{P}$ is derived by minimizing a cost function similar to Graph embedding method (GE)~\cite{YanEtal:07Graph} as follows:

\begin{equation}
\label{c_s}
\mathbf{P} = \min_{\mathbf{P}} \frac{1}{2} \sum_{i,j} ||\mathbf{P} \mathbf{x}_i - \mathbf{P} \mathbf{x}_j ||^2 \mathbf{W}_{ij},
\end{equation} 

where $\mathbf{W}$ is a sparse symmetric $N_{\mathcal{L}} \times N_{\mathcal{L}}$ matrix that represents class similarity relationship and where $\mathbf{W}_{ij} = 1$ if $l_i = l_j$ and $\mathbf{W}_{ij} = -1$ if $l_i \ne l_j$.

Given the transformation $\mathbf{P}$ of the data from LTM, we can find a mapping $\mathbf{Q}: \mathbf{Y} \rightarrow \mathbf{V}$ such that each data instance $\mathbf{y}_i \in \mathbb{R}^n$ with class $l_i \in C$ is mapped to a point $\mathbf{v}_i \in \mathbb{R}^d$ in the neighborhood of the cluster corresponding to the class $l_i$.
In this paper, we provide an approximation solution to find such a mapping. 
The idea is that each data instance $\mathbf{y}_i$ with class label $l \in C$ is mapped to a point $\mathbf{v}_i$ such that $\mathbf{v}_i$ is as close as possible to the median point $\mathbf{\gamma}_l$.
We can find such a linear mapping matrix $\mathbf{Q} \in \mathbb{R}^{d \times n}$ by setting:

\begin{equation}
\label{eqn:ridge1} 
\mathbf{Q} = \arg \min_{\mathbf{Q}} \sum_l \sum_{i:l_i=l} || \mathbf{Q} \mathbf{y}_i  - \mathbf{\gamma}_l ||^2,
\end{equation}

where $\sum_{i:l_i=l}$ denotes the summation over $i$ such that label $l_i = l$.

To avoid overfitting, we add a regularization term $\zeta(\mathbf{Q})$. For example, we can set $\zeta(\mathbf{Q}) = \eta||\mathbf{Q}||_F^2$, in which $\eta$ is the weight and $|| \cdot ||_F$ is the Frobenius norm.
The $\mathbf{P}$ and $\mathbf{Q}$ mapping functions are derived for the purpose of obtaining a low-dimensional representation of the data that separates the populations as much as possible.
However, the classes may not be sufficiently separated.
Consequently, we apply a discrimination method again to determine a mapping function $\mathbf{H}$ to render the classes as separated as possible.


\subsection{Feature Augmentation}
\label{sec:augmentation}
After applying the final discrimination method to transfer the data, we derive new features.
Daum{\'e} III provides a \textit{feature augmentation} method to integrate more information~\cite{Daume:07Frustratingly}.
For each instance, we utilize the feature augmentation method to augment the new features with the original features to represent the instance~\cite{Daume:07Frustratingly}.





We define two feature mapping functions $\phi_{\mathcal{L}}(\mathbf{x}) = [(\mathbf{HP}\mathbf{x})^\top, 0_{n}^\top]^\top$ and $\phi_{\mathcal{S}}(\mathbf{y}) = [(\mathbf{HQ}\mathbf{y})^\top, \mathbf{y}^\top]^\top$ for the data from LTM and SM, respectively.
Here $0_{n}$ denotes zero column vectors of dimensions $n$.
Including zeros in the feature representations ensures that the dimensions of the instances from LTM and SM become the same.
Moreover, the entire data set from LTM has no information on the features from SM.
Therefore, it is reasonable to set zero values ($0_{n}^\top$) in the later $n$ feature dimensions.

\subsection{Proposed Algorithmic Procedure}
As a summary, we give the algorithm of the proposed method:
\begin{enumerate}
 \item Use a projection approach to learn a mapping $\mathbf{P}$ to project data retrieved from LTM to a new $d$ dimensional space in the working memory. All the projected data in the new space are grouped into several clusters corresponding to the classes.
 \item Compute the generalized geometric median of all clusters.
 \item Compute the mapping function $\mathbf{Q}$ for data in the SM via these generalized geometric medians.
 \item Apply $\mathbf{Q}$ to map the SM data for training and new incoming SM data to the new $d$ dimensional space.
 \item Apply a discrimination method again to construct a transformation matrix $\mathbf{H}$ to separate \textit{all} the instances into distinct classes.
 \item Augment the learned features with the original SM features to represent the instance.
 \item With the projected data with known classes from LTM and SM as training data, conventional supervised machine learning methods can be used to learn the model and predict the projected new incoming data from SM. 
\end{enumerate}

\section{Experimental Results}
\label{sec:experiments}

In this section, we conduct several experiments on two benchmark data sets. 
We assume that there is only one data set from LTM and one data set from SM.
Because the upper bound of the number of retained dimensions in LDA is $c-1$ ($c$ is the number of classes), we set the dimension of the new space in the first projection for learning $\mathbf{P}$ and final projection to $c-1$.

For selection of the weight of the regularization term $\eta$, cross-validation is not applicable due to the small number of training instances from the SM domain.
Therefore, we tune the parameters on a predefined range and report the optimal parameter value.
In this paper, we report the results when $\eta = 1$.

We apply a k nearest neighbor classifier (kNN, k = $5$) and a Support Vector Machine with a Radial Basis Function kernel (RBF SVM) to train the final classifiers on the data sets.
The training instances are derived from both LTM and SM. 
For the baseline approach, we train the classifier on the original labelled instances from SM and use it to predict the test instances. 
We randomly sample the training instances ten times and report the average accuracy over the ten rounds of experiments.
In this paper, we conduct experiments in different settings in the two data sets and report the results.

\subsection{Data Sets}
\subsubsection{Object Recognition Data Set}

The first data set\footnote{Visit http://vision.cs.uml.edu/adaptation.html for more details.} contains $4,652$ images from $31$ categories\footnote{These $31$ categories are: backpack, bike, bike helmet, bookcase, bottle, calculator, computer, desk chair, desk lamp, file cabinet, headphones, keyboard, laptop, letter tray, mobile phone, monitor, mouse, mug, notebook, pen, phone, printer, projector, puncher, ring binder, ruler, scissors, speaker, stapler, tape, and trash can} originating from the following three domains: \texttt{Amazon} (images downloaded from an online retail website), \texttt{dslr} (high-resolution images taken from a digital DLR camera) and \texttt{webcam} (low-resolution images taken from a web camera)~\cite{SaenkoEtal:10Adapting}.
SURF features are extracted for all the images.

The \texttt{Amazon} and \texttt{webcam} images are used as data retrieved from LTM.
We randomly select twenty and eight training images per category for the \texttt{Amazon} and \texttt{webcam} data sets, respectively. 
We then randomly select $k$ training images per category for the \texttt{dslr} data set as SM data, where $k = 3, 4, 5,$ and $6$.
The remaining images are used for testing. 
Table~\ref{tbl:saenkol_dataset} shows a summary of the data set.

\begin{table}[h]
\scriptsize
\centering
\caption{Summarization of the object recognition data set.}
\label{tbl:saenkol_dataset}
\begin{tabular}{@{}cccccc@{}}
\toprule
 &\multirow{2}{*}{Data set} & \# total  & \multirow{2}{*}{\# dim} & \# training data \\
  &   &   instances &        & instances per class  \\ \midrule     
  \multirow{2}{*}{LTM}&\texttt{Amazon} & $2813$ & $800$ & $20$  \\
  & \texttt{webcam} &  $795$ & $800$ & $8$ \\ \cmidrule(r){1-5}
  SM & \texttt{dslr}   & $498$  & $600$ & $3/4/5/6$  \\ \bottomrule
\end{tabular}%
\end{table}

\subsubsection{Text Categorization Data Set} 
The second data set is a subset of the Reuters RCV1/RCV2 collections~\cite{AminiEtal:09Learning}.
It contains newswire articles written in \texttt{English}, \texttt{French}, \texttt{German}, \texttt{Italian}, and \texttt{Spanish}.
There are six classes for these articles: \texttt{C15}, \texttt{CCAT}, \texttt{E21}, \texttt{ECAT}, \texttt{GCAT}, and \texttt{M11}.

We take \texttt{Spanish} articles as SM data and articles written in the other four languages as individual data sets retrieved from LTM.
For each class, we randomly sample one hundred training instances from the data sets in LTM and $k$ training instances from SM data, where $k = 5, 7, 10, 15,$ and $20$.
We also perform a Principal Components Analysis (PCA) with $60\%$ energy preserved on the TF-IDF features.
The remaining instances from SM data are used as the test instances.
Table~\ref{tbl:reuters_dataset} shows a summary of the data set, and Table~\ref{tbl:reuters_dataset_class} shows the distribution of classes in the data set.

\begin{table}[h]
\begin{minipage}[b]{.50\textwidth}
\scriptsize
\centering
\caption{Summarization of the text categorization data set}
\label{tbl:reuters_dataset}
\begin{tabular}[t]{@{}ccccc@{}} \toprule
 & \multirow{2}{*}{Data set} & \# dim &\# total   & \# training data \\
  &   & after PCA    &   instances & instances per class  \\ \midrule 
\multirow{4}{*}{LTM} & \texttt{English} & $1,131$ & $18,758$  & \multirow{4}{*}{$100$}\\
    & \texttt{French} & $1,230$ &   $26,648$   &  \\
    & \texttt{German} & $1,417$ &  $29,953$   &  \\ 
    & \texttt{Italian} & $1,041$ &  $24,039$     &  \\ \cmidrule(r){1-5}
 SM & \texttt{Spanish} & $807$ &  $12,342$ & $5/7/10/15/20$  \\ \bottomrule
\end{tabular}%
\end{minipage}\qquad
\begin{minipage}[b]{.50\textwidth}
\scriptsize
\centering
\caption{Distribution of classes in the text categorization data set}
\label{tbl:reuters_dataset_class}
\begin{tabular}[t]{@{}ccc@{}} \toprule
Class & Size(all languages) & ($\%$) \\ 
\midrule
\texttt{C15} & $18,816$ &   $16.84$ \\
\texttt{CCAT} & $21,426$ &  $19.17$ \\ 
\texttt{E21} & $13,701$ &  $12.26$  \\
\texttt{ECAT} & $19,198$ &  $17.18$ \\ 
\texttt{GCAT} & $19,178$ & $17.16$ \\
\texttt{M11} & $19,421$ & $17.39$ \\ \bottomrule
\end{tabular}%
\end{minipage}
\end{table}

\subsection{Comparative Studies}
\subsubsection{Object Recognition Data Set}

Table~\ref{tbl:object_tr_da_dataset} shows the performance of the baseline approach and the CDM approach in different settings on the object recognition dataset.
The results show that the CDM approach using a kNN classifier both with and without augmentation have outstanding performance in the object recognition dataset. 
In the case of the RBF SVM classifier, although the learned features are not good enough for training a prediction model, augmenting the learned features with the original features from SM improves the performance. 

\begin{table}[h]
\scriptsize
\centering
\caption{Means and standard deviations of classification accuracies $(\%)$ of the baseline and the CDM approaches on the object recognition dataset by using $3$ labeled training samples per class from the SM domain \texttt{dslr}. The CDM approach shows the results in different settings.}
\label{tbl:object_tr_da_dataset}
\begin{tabular}{@{}ccccc@{}} \toprule
 \multirow{2}{*}{LTM} & \multirow{2}{*}{Classifier} & \multirow{2}{*}{Baseline}& Without & With \\  
 &  & & augmentation & augmentation \\ \midrule
\multirow{2}{*}{\texttt{Amazon}}& kNN & $32.6 \pm 2.3$ & $45.4 \pm 2.2$ &  $44.0 \pm 2.3$ \\ 
& SVM  & $50.6 \pm 3.0$ & $39.4 \pm 3.4$ & $54.2 \pm 3.0$ \\ \midrule  
\multirow{2}{*}{\texttt{webcam}}& kNN & $32.6 \pm 2.3$ & $42.9 \pm 2.7$ & $42.6 \pm 2.8$ \\  	    
& SVM  & $50.6 \pm 3.0$ & $42.3 \pm 1.7$ & $56.2 \pm 2.5$ \\ \bottomrule
\end{tabular}
\end{table}

\subsubsection{Text Categorization Data Set} 

Table~\ref{tbl:reuters_tr_da_dataset} shows the performance of both the baseline approach and the CDM approach in different settings on the text categorization dataset.
Both the kNN and RBF SVM classifiers with and without augmentation outperform the baseline approach.

\begin{table}[h]
\scriptsize
\centering
\caption{Means and standard deviations of classification accuracies $(\%)$ of the baseline and CDM approaches on the text categorization dataset by using $20$ labeled training samples per class from the SM domain Spanish. The CDM approach shows the results in different settings.}
\label{tbl:reuters_tr_da_dataset}
\begin{tabular}{@{}ccccc@{}} \toprule
 \multirow{2}{*}{LTM} & \multirow{2}{*}{Classifier} & \multirow{2}{*}{Baseline}& Without & With  \\  
 &  & & augmentation & augmentation \\ \midrule
\multirow{2}{*}{\texttt{English}}& kNN & $28.3 \pm 11.4$ & $66.5 \pm 3.9$ & $66.5 \pm 3.9$ \\                                                                                                                               
& SVM  & $44.2 \pm 3.3$ & $65.3 \pm 5.0$ & $61.6 \pm 3.9$ \\ \midrule                                                               
\multirow{2}{*}{\texttt{French}}& kNN & $28.3 \pm 11.4$ & $67.4 \pm 4.5$ & $67.4 \pm 4.5$ \\                                                                
& SVM  & $44.2 \pm 3.3$ & $66.9 \pm 6.1$ & $61.4 \pm 3.7$ \\ \midrule
\multirow{2}{*}{\texttt{German}}& kNN & $28.3 \pm 11.4$ & $67.4 \pm 3.2$ & $67.4 \pm 3.2$ \\
& SVM  & $44.2 \pm 3.3$ & $69.3 \pm 3.9$ & $60.6 \pm 3.8$ \\ \midrule                                                                
\multirow{2}{*}{\texttt{Italian}}& kNN & $28.3 \pm 11.4$ & $66.6 \pm 2.8$ & $66.6 \pm 2.8$ \\                                                                 
& SVM  & $44.2 \pm 3.3$ & $67.2 \pm 3.7$ & $60.8 \pm 3.3$ \\ \bottomrule
\end{tabular}
\end{table}

\subsection{Influence of the Number of Training Samples per Class from SM domain}

\begin{figure}[h]
\centering
\begin{subfigure}[t]{.5\textwidth}
\centering
  \includegraphics[width=0.9\linewidth]{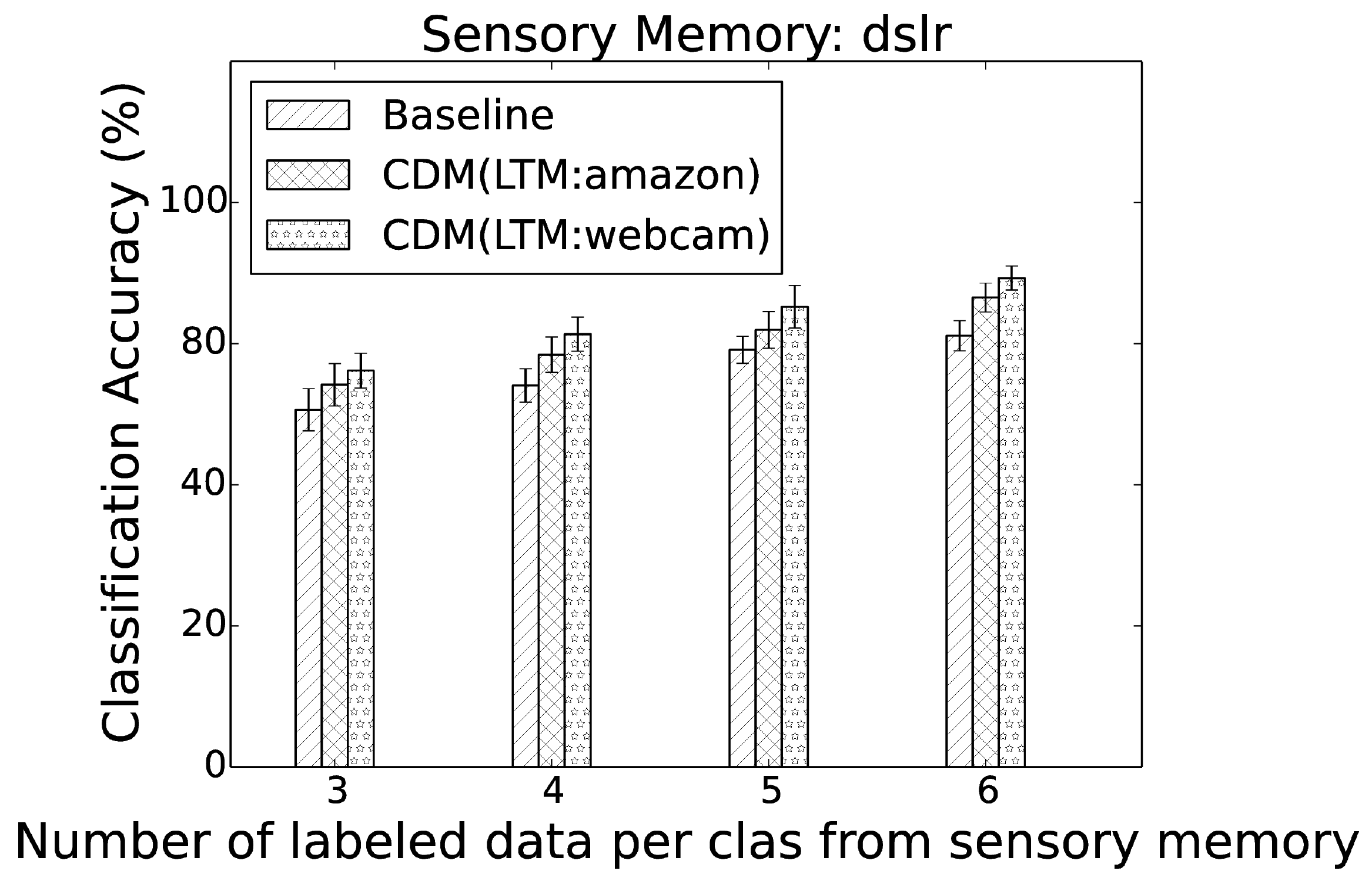}
  \caption{The object recognition data set}
  \label{fig:object_k}
\end{subfigure}%
\begin{subfigure}[t]{.5\textwidth}
\centering
\includegraphics[width=0.9\linewidth]{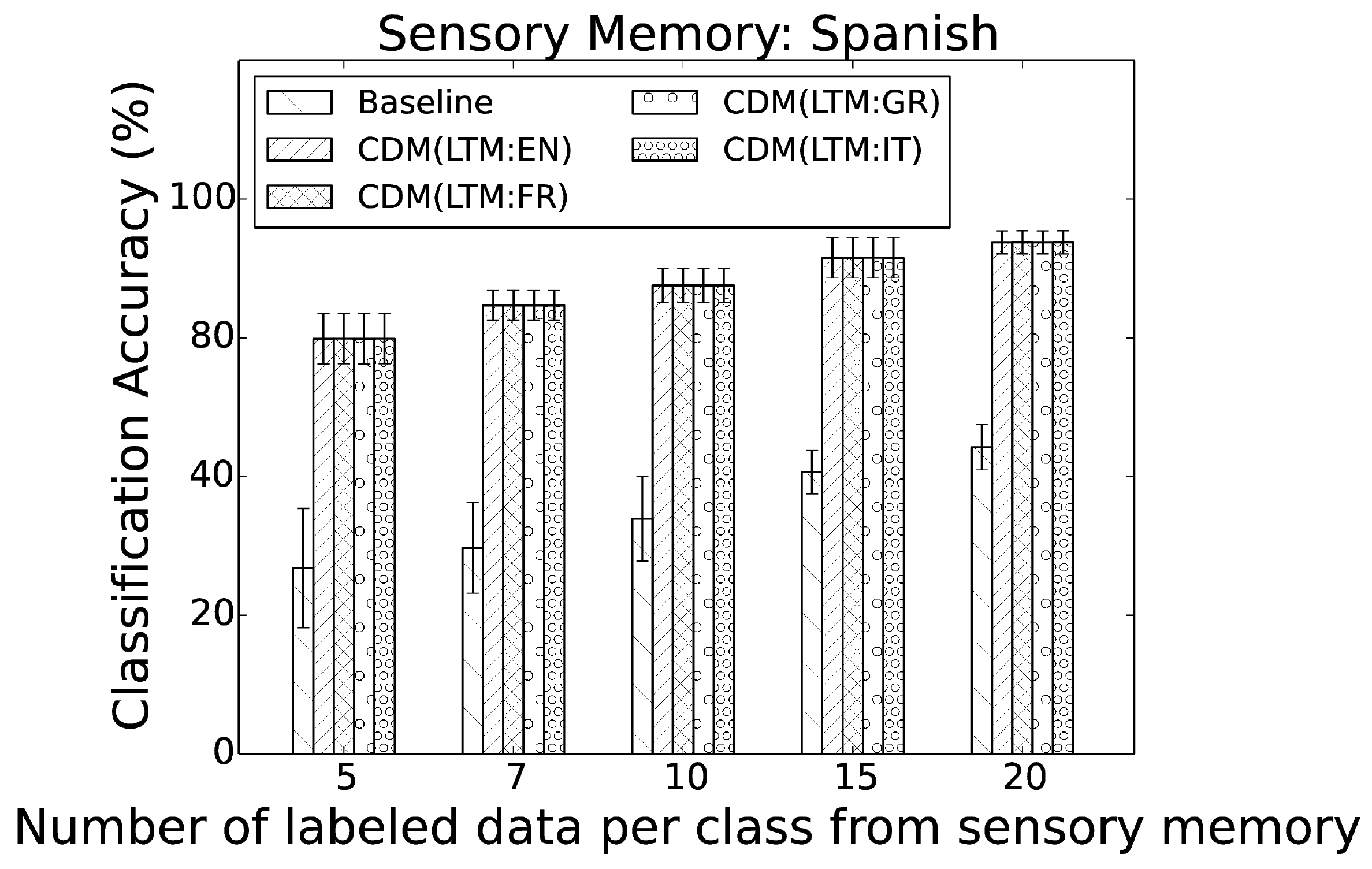}
  \caption{The text categorization data set}
  \label{fig:text_k}
\end{subfigure}%
\caption{Classification accuracies of the baseline approach and CDM approach with respect to different numbers of SM training samples per class on the object recognition data set and the text categorization data set. In the case of CDM, different hatching patterns correspond to different LTM domains.}
\label{fig:k}
\end{figure}

Figure~\ref{fig:k} shows the accuracy of baseline and the proposed CDM approach with respect to the number of training samples per class for each data set from SM.
For both the CDM and baseline approaches, we use the RBF SVM as the classifier.
As shown in Fig.~\ref{fig:object_k} and Fig.~\ref{fig:text_k}, the accuracies of the baseline approach and CDM increase when we use a larger $k$.

\section{Conclusions}
\label{sec:conclusion}
We proposed a computational model called CDM for the cognitive rapid learning problem.
CDM has the following excellent properties:
\begin{inparaenum}[\itshape a\upshape)]
\item CDM is elegant, intuitive, and efficient; and,
\item CDM has a complete theoretical architecture. 
\end{inparaenum}
The experimental results show that we can find at least one hypothesis or classifier that satisfies our hypothesis for rapid learning.
One-shot learning or rapid learning is still a work in progress.
From the CDM experience, we think that cognitive scientists should be required to pay attention to the role of information retrieved from LTM in rapid learning.

From the experimental results, we found that the feature augmentation did not always guarantee performance improvement.
It is worth investigating this issue in the future.    

In CDM, because we want to create a common metric space for LTM and SM, we proposed \emph{global} metric learning in the working memory; unfortunately, this results in a CDM limitation.
CDM attempts to find mapping matrices that minimize the sum of \emph{all} pairwise distances between data points in the same class and maximize the sum of \emph{all} pairwise distances between data points in different classes.
As such, this implicitly assumes that classes form a single compact connected set.
If the data instances are in highly multi-modal class distributions, the cost function will be penalized.
Therefore, developing a discrimination approach that focuses on \emph{local} neighborhoods for the problem is worth studying in future research.



\bibliographystyle{plain}
\bibliography{nips2016}

\end{document}